\pdfoutput=1

\documentclass[11pt]{article}

\usepackage[preprint]{acl}

\usepackage{times}
\usepackage{latexsym}
\usepackage{booktabs}
\usepackage{array}
\usepackage{url}
\usepackage[breaklinks]{hyperref}
\usepackage{breakurl}
\usepackage[breaklinks]{hyperref}

\usepackage[T1]{fontenc}

\usepackage[utf8]{inputenc}

\usepackage{microtype}

\usepackage{inconsolata}

\usepackage{graphicx}

\title{Mathematical Computation and Reasoning Errors by Large Language Models\thanks{This paper has been officially accepted for presentation at the Artificial Intelligence in Measurement and Education Conference (AIME-Con) 2025, held in Pittsburgh, Pennsylvania, United States, from October 27 to 29, 2025.}}

\author{Liang Zhang\\
Institute for Intelligent Systems \\ University of Memphis, Memphis, TN, USA \\
  AI4STEM Education Center \\ University of Georgia, Athens, GA, USA \\
  \texttt{lzhang13@memphis.edu} \\\And
  Edith Aurora Graf \thanks{This research was conducted while the second author was at Educational Testing Service. }\\
 \\ Lawrenceville, NJ, USA \\
  \texttt{eag2718@gmail.com} \\}

\begin{document}
\maketitle
\begin{abstract}
Large Language Models (LLMs) are increasingly utilized in AI-driven educational instruction and assessment, particularly within mathematics education. The capability of LLMs to generate accurate answers and detailed solutions for math problem-solving tasks is foundational for ensuring reliable and precise feedback and assessment in math education practices. Our study focuses on evaluating the accuracy of four LLMs (OpenAI GPT-4o and o1, DeepSeek-V3 and DeepSeek-R1) solving three categories of math tasks, including arithmetic, algebra, and number theory, and identifies step-level reasoning errors within their solutions. Instead of relying on standard benchmarks, we intentionally build math tasks (via item models) that are challenging for LLMs and prone to errors. The accuracy of final answers and the presence of errors in individual solution steps were systematically analyzed and coded. Both single-agent and dual-agent configurations were tested. It is observed that the reasoning-enhanced OpenAI o1 model consistently achieved higher or nearly perfect accuracy across all three math task categories. Analysis of errors revealed that procedural slips were the most frequent and significantly impacted overall performance, while conceptual misunderstandings were less frequent. Deploying dual-agent configurations substantially improved overall performance. These findings offer actionable insights into enhancing LLM performance and underscore effective strategies for integrating LLMs into mathematics education, thereby advancing AI-driven instructional practices and assessment precision. 
\end{abstract}

\section{Introduction}

Large Language Models (LLMs) have significantly impacted mathematical instruction and assessment. Educational platforms are increasingly integrating LLMs to enhance teaching and evaluation methods. For instance, Khan Academy utilized the LLM-powered tool \textit{Khanmigo} for Socratic-style math assistance \cite{anand2023khan}. Coursera uses LLMs to streamline assessment creation, automate grading, and offer personalized feedback \cite{maggioncalda2024coursera}. Quizlet’s \textit{Q-Chat} integrates LLM-based conversational AI to dynamically adapt question difficulty levels and deliver guided hints \cite{quizlet2025qchat}. Beyond merely producing final answers, LLMs excel at clearly articulating intermediate computational steps and reasoning processes, significantly enhancing their value in mathematics education contexts \cite{gupta2025beyond}. These capabilities of LLMs facilitate personalized tutoring, interactive problem-solving, and real-time feedback, significantly reduce grading workloads and ensure consistent evaluations in mathematics education.

The capability of LLMs to produce accurate answers and detailed, step-by-step solutions in math problem-solving is foundational for reliable assessment and precise feedback in mathematics education \cite{gupta2025beyond,jin2025investigating}. Specifically, LLM-based automated assessment involves evaluating granular math skills through step-level grading of student solutions \cite{jin2025investigating}, performing automatic step-level corrections \cite{li2025teaching}, and providing targeted instructional hints \cite{tonga2025simulating}. However, this raises an important question: if LLMs cannot reliably produce correct answers or accurately solve math problems, can their outputs still be considered effective and trustworthy for instructional guidance and learner assessment? This motivated our idea to systematically test the capability of LLMs to accurately solve diverse math tasks, and subsequently extend their applicability toward realistic assessment and instructional scenarios. While state-of-the-art LLMs have demonstrated high accuracy on various math benchmarks, benchmark success alone does not present a comprehensive picture. Performance significantly declines on certain math tasks like fundamental numerical understanding and basic computational problems \cite{yang2024number, boye2025large,petrov2025proof}. Current LLM computation and reasoning processes remain prone to calculation mistakes (e.g., arithmetic slips, algebraic simplification errors) and logical reasoning errors (e.g., invalid inference steps, omission of necessary procedural steps, and self-contradictory reasoning) \cite{li2024evaluating,roy2025can}. These persistent errors significantly limit the reliability and efficacy of LLM outputs for instructional feedback and learner assessment purposes. 

In this study, we specifically examine the capabilities and limitations of LLMs in math problem-solving, focusing on assessing the accuracy of generated answers and identifying errors within solution steps. Instead of using benchmarks, we built math problems in arithmetic, algebra, and number theory to evaluate LLMs’ proficieny in math computation and reasoning. We explored four distinct LLMs, two base models GPT-4o \cite{hurst2024gpt} and DeepSeek-V3 \cite{liu2024deepseek} and two reasoning-enhanced models OpenAI o1 \cite{jaech2024openai} and DeepSeek-R1 \cite{guo2025deepseek}, across three math problem-solving tasks, including arithmetic, algebra, and number theory. Two interaction paradigms are considered: (i) a single-agent setting in which one model works through each math task step-by-step, and (ii) a dual-agent setting that lets two peer LLMs chat, cross-validate, and refine their reasoning, echoing recent advances in collaborative intelligence \cite{zhang2025exploring,latif2024systematic,zhang2025exploring_aied}. Every solution was decomposed into granular steps and coded with an expert-verified rubric, which enabled us to quantify step-level accuracy and localize procedural or conceptual errors. Our investigation is guided by two \textbf{R}esearch \textbf{Q}uestions: \textbf{Q1:} How accurately do LLMs generate final answers to math problems? 
\textbf{Q2:} What recurring error patterns (procedural, conceptual, or logical) emerge in their step-level solutions?

This study will provide researchers and practitioners with precise insight into where LLMs excel and where they falter in math computation and reasoning. We also provide a rubric that can be used to evaluate the accuracy of LLM-based solutions, and to identify the nature of errors when they occur. Our study demonstrates that knowledgeable LLMs have potential to reliably support math instruction and assessment, and we offer actionable guidance for their effective use.

\section{Related Works}


LLMs have made notable strides in solving math problems, yet they frequently struggle with precise computations and multi-step numerical reasoning tasks \cite{wolfram2023wolfram,li2024evaluating}. Frieder et al. \cite{frieder2023mathematical} provided a detailed evaluation demonstrating that GPT-4 (the 2023 version of ChatGPT) effectively handles many undergraduate-level questions but exhibits significant difficulties when confronted with graduate-level math challenges, particularly in proof-based tasks and complex symbolic computations. Yang et al. \cite{yang2024number} found that LLMs frequently make surprising mistakes in basic numerical understanding and processing tasks. Some studies \cite{li2024evaluating,pan2025lemma} have identified recurring error patterns in math reasoning by LLMs, including calculation errors, counting errors, formula confusion, question misinterpretation, missing solution steps, conceptual confusion, and nonsensical outputs, among others. Numerous additional studies have consistently reported similar challenges, emphasizing the ongoing limitations faced by LLMs in performing math tasks \cite{arkoudas2023gpt,wolfram2023wolfram,wang2023scibench,zhang2024careful,mcleish2024transformers}. 

Mitigating math problem-solving errors and boosting LLM performance is a multifaceted endeavor. Technical advances center on modular reasoning strategies, such as Chain-of-Thought and Program-of-Thought prompting \cite{wei2022chain, chen2022program}, alongside fine-tuning and math-specific training regimens \cite{zhang2023interpretable, ahn2024large}, novel architectures that integrate external tools or structured reasoning modules, and rigorous evaluations that precisely expose weaknesses. On the usability side, carefully designed prompts, curriculum-aligned task sequencing, interactive dialogue, and built-in self-checking routines can substantially reduce errors in real-world use.



\section{Methods}


\textbf{Dataset.} For the dataset, we used the problem categories and instances developed for \citet{graf2025math}. Three distinct types of math tasks were utilized to evaluate the performance of LLMs: (1) multiplying two 5-digit numbers; (2) solving algebraic word problems involving quadratic equations; and (3) finding solutions to Diophantine equations. See the math dataset in the GitHub repository: 
\url{https://github.com/LiangZhang2017/math_number_computing}. In our approach, we leveraged item models \cite{bejar2002generative,laduca1986item} and automated item generation (AIG) \cite{embretson1999generating,gierl2013automatic,irvine2002item}. An item model is a set of items that share a common structure, defined through the use of variables and constraints. \citet{mirzadeh2024gsmsymbolicunderstandinglimitationsmathematical} used a model-based approach to assess LLM math reasoning by generating template-based variants of existing tasks. These new instances avoided leakage, and performance often differed from the original tasks—especially when variations involved numeric values or complexity rather than names. 
Each problem category was represented as an item modeland used to generate 10 instances. The item models  were defined as follows: (1) Item model 1 involves finding the product of two 5-digit whole numbers. (2) Item model 2 involves finding two distinct two-digit whole numbers with a given sum and a given product, where the sum is less than or equal to 100, and neither number is divisible by 10. (3) Item model 3 involves finding a pair of positive integers \textit{x, y} that satisfy an equation of the form \(p_1x^a=p_2y^b\), where \(p_1\) and \(p_2\) are distinct primes such that each is less than or equal to 11. The exponents \textit{a, b} are relatively prime and each is less than or equal to 9. 


\textbf{LLM Models Setup.} We defined two scenarios for configuring LLMs as agents to solve math problems: single-agent and dual-agent setups. Both scenarios are designed to elicit comprehensive, step-level solutions as well as accurate final answers. In the single-agent scenario, individual LLMs, including OpenAI GPT-4o, DeepSeek-V3, OpenAI o1, and DeepSeek-R1, are configured as math problem-solving assistant agents that independently perform math problem-solving tasks. In the dual-agent scenario, two base LLM models (OpenAI GPT-4o and DeepSeek-V3) collaborate as peer agents through interactive, chat-based discussions, exchanging ideas and jointly deriving solutions. Each setup was repeated in three independent runs to ensure reliability. 

\textbf{Evaluations.} For instances of the first two item models, the answer key is either a single value (Item Model 1) or, assuming the two integers are interchangeable, a single pair of values (Item Model 2). For instances from Item Model 3, however, there are infinitely many pairs \(x\), \(y\) that satisfy the given equation. Since only one pair  \(x\), \(y\) is requested however, evaluating correctness can be accomplished by substituting the provided values for \(x\), \(y\) into the given equation--if this yields a true result, the response is correct, otherwise, it is incorrect. Since solutions are always possible, any response that states there are no solutions is incorrect. To evaluate the solutions, we used a structured coding process: (1) each solution was segmented into discrete, logical steps. (2) each step was labeled according to a predefined rubric detailed in Table~\ref{tab:solutions_coding_rubric}, categorizing step labels as CC (Conditionally Correct), PE (Procedural Error), CE (Conceptual Error), or IE (Impasse Error). The labeling was performed by the o1 LLM model, followed by verification from human experts. We applied a conditional scoring approach to avoid penalizing LLMs for errors made in earlier solution steps. Analysis of labeling patterns will be reported separately. 

\begin{table}[ht]
    \centering
    \scriptsize
    \caption{Math Problem Solution Coding Rubric.}
    \label{tab:solutions_coding_rubric}
    \renewcommand{\arraystretch}{1.4}
    \begin{tabular}{>{\raggedright\arraybackslash}p{0.15\textwidth}p{0.27\textwidth}}
        \toprule
        \textbf{Step Code} & \textbf{Definition} \\
        \midrule
        Conditionally Correct (CC) & A step that demonstrates procedural and conceptual accuracy, controlling for any errors that may have occurred on previous steps. \\
        Procedural Error (PE) & A step that contains one or more transcription errors, arithmetic mistakes, or symbolic manipulation errors, but without underlying conceptual misunderstanding. \\
        Conceptual Error (CE) & A step demonstrating one or more incorrect applications or misunderstandings of relevant math concepts or principles. It may include misunderstanding the problem, representing it incorrectly, or committing reasoning errors between steps.\\
        Impasse Error (IE) & A step where the solver is unable to proceed further logically or mathematically, indicating a critical gap or blockage in problem-solving understanding. \\
        \bottomrule
    \end{tabular}
\end{table} 

\section{Results and Discussion}

The preliminary results presented below include systematic testing of LLMs' performance across three distinct types of math problems, labeling outcomes that identify solution errors across these problem types, and an initial exploration of collaborative LLM-based agents for math tasks. 

\begin{figure}[h!t]
\centering
\includegraphics[width=3in]{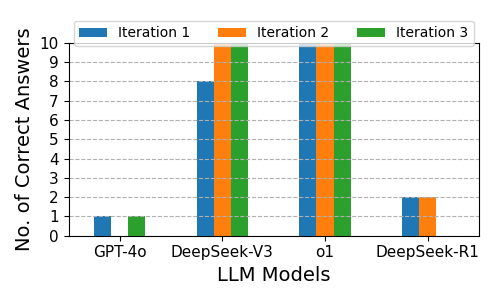}
\caption{Correctness Across Three Iterations for the Multiplying Two 5-digit Numbers.}
\label{fig:product_comparison}
\end{figure}

\begin{figure}[h!t]
\centering
\includegraphics[width=3in]{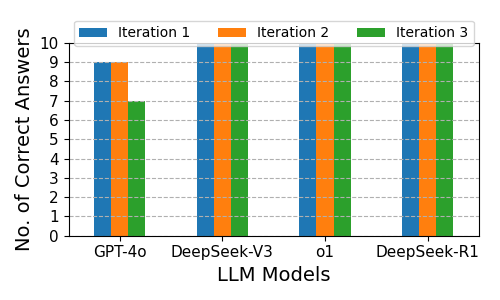}
\caption{Correctness Across Three Iterations for Solving Algebraic Word Problems Involving Quadratic Equations.}
\label{fig:algebraic}
\end{figure}

\begin{figure}[h!t]
\centering
\includegraphics[width=3in]{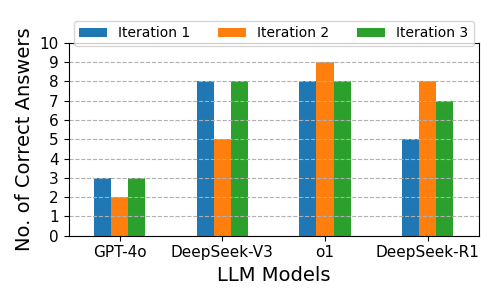}
\caption{Correctness Across Three Iterations for Solving Diophantine equations.}
\label{fig:diophantine}
\end{figure}

\textbf{Accuracy of Final Answers from Single-Agent. } Figure~\ref{fig:product_comparison} presents the performance of four LLMs on a math problems involving the multiplication of two 5-digit numbers. GPT-4o exhibited the lowest performance, with only two correct answers overall. DeepSeek-V3 started strong (8/10 correct) and quickly achieved perfect accuracy in the subsequent iterations (28/30 total). The o1 model demonstrated flawless accuracy from the outset, solving all problems correctly across all three iterations. DeepSeek-R1 achieved only four correct answers across all three runs combined. Interestingly, while the reasoning-enhanced o1 model significantly outperformed its base counterpart GPT-4o, this was not the case with DeepSeek-R1 relative to DeepSeek-V3. We found that DeepSeek-R1 struggled substantially on our proposed tasks. Upon examining its detailed solutions, the model appeared to exhibit an ``overthinking'' phenomenon (akin to ``spinning wheels''), characterized by excessive reflection on intermediate reasoning steps, causing it to overlook critical components necessary for accurate solutions. This outcome of DeepSeek-R1 deviates from performance reported in prior benchmark evaluations \cite{guo2025deepseek}. As shown in Figure~\ref{fig:algebraic}, we evaluated the number of correct answers provided by each model for algebraic word problems involving quadratic equations. GPT-4o demonstrated moderate performance, correctly solving 9 out of 10 problems in Iterations 1 and 2, but experiencing a slight drop to 7 correct solutions in Iteration 3. In contrast, DeepSeek-V3, o1, and DeepSeek-R1 consistently achieved perfect accuracy, correctly solving all 10 problems across each of the three iterations. Figure~\ref{fig:diophantine} presents the accuracy results, showing the performance ranking as follows: o1 (25/30) > DeepSeek-V3 (21/30) > DeepSeek-R1 (20/30) > GPT-4o (8/30). The advanced o1 model clearly outperformed its base counterpart GPT-4o; however, this was not the case within the DeepSeek series. 

\begin{figure}[h!t]
\centering
\includegraphics[width=3in]{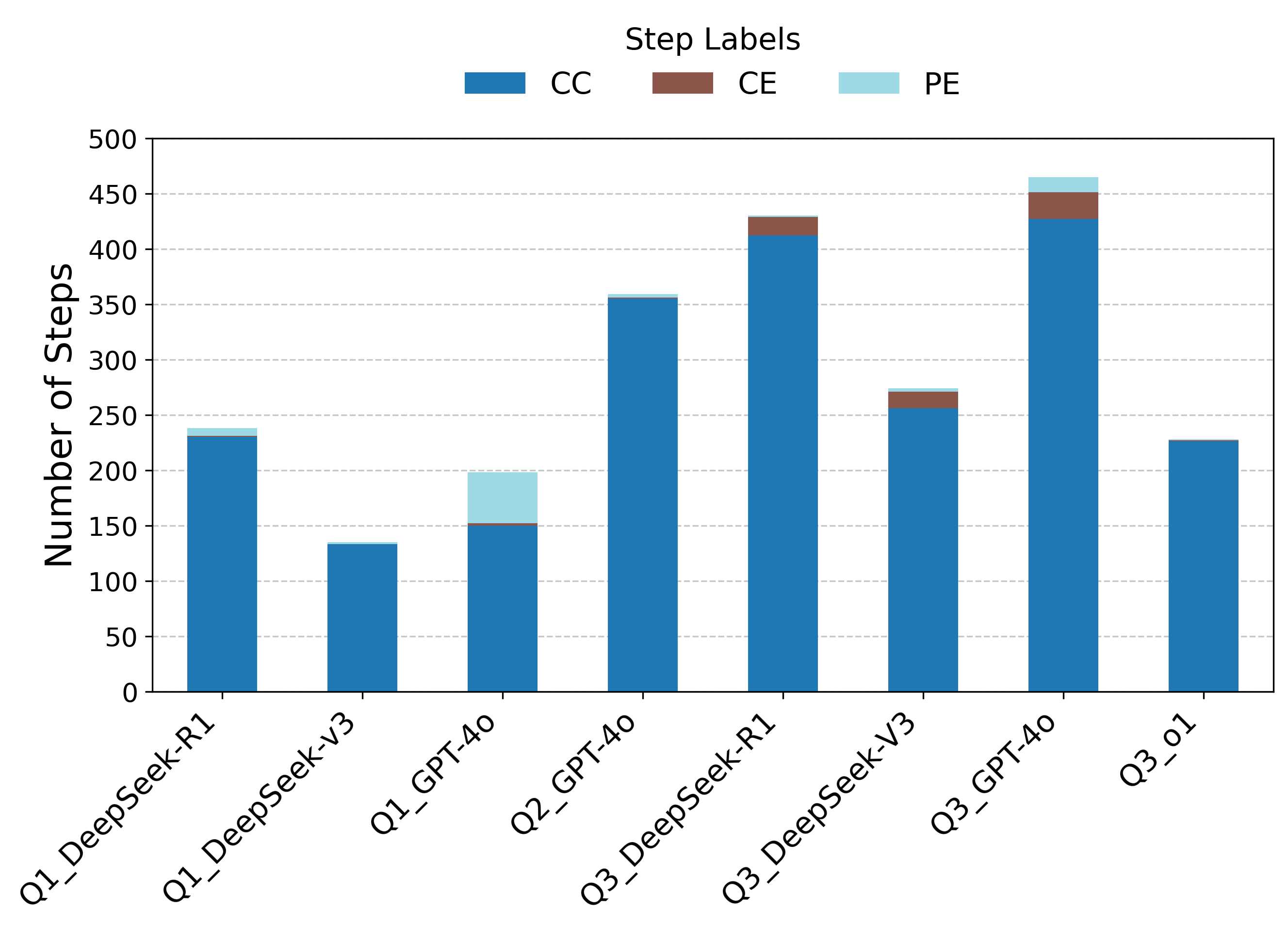}
\caption{Frequencies of Step Labels in Math Tasks (Where LLMs Stumble). Note: Q1= multiplying two 5-digit numbers; Q2= solving algebraic word problems, Q3= finding solutions to Diophantine equations. We only labeled sets of math problems in which the LLMs produced incorrect final answers, excluding those with 100\% correctness.}
\label{fig:llm_labeling}
\end{figure}

\textbf{Evaluations by LLM-based Labeling in the Single-Agent Scenario.}  
The labeled steps in math tasks across the three iterations of the single-agent scenario are shown in Figure~\ref{fig:llm_labeling}. We specifically selected these math tasks due to their tendency to highlight significant errors, reflecting notably lower accuracy for some LLM models. Among all tasks, the ``CC'' label consistently occurs with the highest frequency across models. However, the presence of ``CE'' labels, notably observed for DeepSeek-R1 (Q3), DeepSeek-V3 (Q3) and GPT-4o (Q3), indicates gaps in understanding fundamental math concepts and principles necessary for accurate solutions, potentially explaining their reduced performance. GPT-4o (Q1) and GPT-4o (Q3) demonstrate the most frequent ``PE'' occurrences, significantly impacting its overall performance (see Figure~\ref{fig:product_comparison}). Conversely, DeepSeek-V3 (Q1) and GPT-4o (Q2) and o1 (Q3) exhibit no clear conceptual misunderstandings or procedural errors, consequently achieving the highest overall accuracy across the math tasks. In these cases, some incorrect final answers occurred despite no clearly identifiable errors  (a phenomenon consistent with our experience using LLMs). A straightforward explanation is that LLMs inherently rely on token prediction rather than explicit numerical computation, rendering them vulnerable to subtle numerical inaccuracies. Further research is necessary to better understand this behavior. Although DeepSeek-R1 (Q1) mostly exhibits minor procedural errors ("PE") in technically correct steps, its overly complex and inefficient reasoning significantly impedes achieving correct final answers. 

As a case study, an expert human coder verified the automated labels produced by GPT-4o and o1 on 70 solution steps drawn from first-iteration problems on the multiplication of two five-digit numbers. We used a verification process rather than an independent coding procedure due to time constraints; however the expert evaluated each step using the rubric in Table~\ref{tab:solutions_coding_rubric}.  Cohen’s $\kappa$ shows that GPT-4o achieves only \emph{fair} agreement with the human coder ($\kappa = 0.366$), whereas o1 attains \emph{substantial} agreement ($\kappa = 0.737$), nearly doubling reliability. These results suggest that LLMs with stronger math competence, such as o1, yield more dependable step-level annotations, reinforcing their suitability for automated formative assessment. 

\begin{figure}[h!t]
\centering
\includegraphics[width=3in]{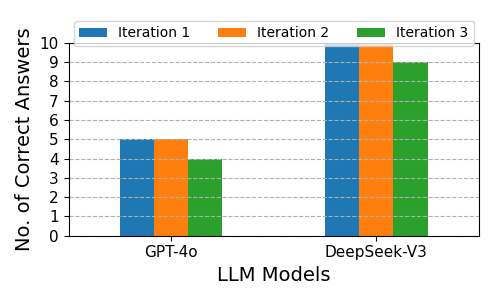}
\caption{Dual-agent Correctness Across Three Iterations for Multiplying Two 5-digit Numbers.}
\label{fig:dual_q1}
\end{figure}

\begin{figure}[h!t]
\centering
\includegraphics[width=3in]{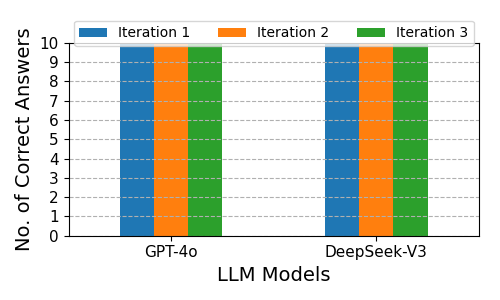} 
\caption{Dual-agent Correctness Across Three Iterations for Solving Algebraic Word Problems Involving
Quadratic Equations.}
\label{fig:dual_q2}
\end{figure}

\begin{figure}[h!t]
\centering
\includegraphics[width=3in]{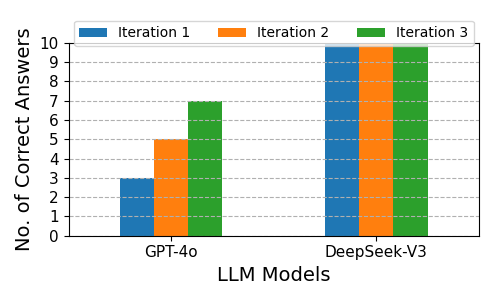} 
\caption{Dual-agent Correctness Across Three Iterations for Solving Diophantine equations.}
\label{fig:dual_q3}
\end{figure}

\textbf{Accuracy of Final Answers from Dual-Agent Collaboration.} Figure~\ref{fig:dual_q1} presents performance results for the dual-agent scenario in solving problems involving the multiplication of two 5-digit numbers. The dual-agent configuration with GPT-4o significantly outperformed the single-agent setup, correctly answering 14 out of 30 questions compared to only 2 out of 32 questions in the single-agent scenario. Figure~\ref{fig:dual_q2} illustrates that both LLM models in the dual-agent scenario achieved perfect accuracy on the quadratic equations questions, surpassing the performance of GPT-4o operating individually as a single agent, which correctly answered only 25 out of 30 problems. Figure~\ref{fig:dual_q3} demonstrates improved accuracy in dual-agent scenarios compared to single-agent setups on the Diophantine equations questions: GPT-4o improved from 8 out of 30 to 15 out of 30, while DeepSeek-V3 notably increased from 21 out of 30 to a perfect 30 out of 30. These results align with findings from Zhang et al.'s study \cite{zhang2025exploring,zhang2025exploring_aied}, highlighting that dual-agent collaboration among LLMs can replicate key benefits of human collaboration. Specifically, collaboration in dual-agent scenarios enhances efficiency by enabling two LLM-based agents to share diverse perspectives, cross-validate solutions, and foster emergent reasoning \cite{chen2023reconcile,liang2024encouragingdivergentthinkinglarge}. Such collaborative mechanisms hold promise for future improvements in math assessment.

\section{Future Work}


Future work should include more detailed labeling to better understand solution errors. For instance, in multiplication problems, errors often occurred in the final step, incorrectly summing partial products. Prompt revisions explicitly instructing step-by-step calculations could reduce such errors. Another promising direction is integrating third-party tools including calculators, spreadsheets, or computer algebra systems to handle computations, with LLMs providing reasoning and explanations. This raises the question of how to decide when to delegate tasks to external tools. Developing fine-tuning methods to improve LLM–tool integration could further enhance accuracy. Assuming math problem-solving performance can be improved through prompt revision or integrating LLMs with third-party tools, a critical question remains: Can such systems effectively support instruction and assessment? Beyond correctness, effective instructional use requires pedagogically sound approaches, and effective assessment demands accurate identification of genuine understanding. Addressing these questions is essential for practical classroom implementation and represents an important next research step. Multi-agent approaches to math problem-solving like the one in this study, which leverage collaborative thinking and collective intelligence \cite{zhang2025exploring,latif2024systematic}, should be further explored. Finally, our findings indicate that stronger performance on final math answers tends to correlate with higher accuracy in step-level assessment. Future studies should investigate the mechanisms underlying this relationship. We also see substantial value in examining how these insights can inform the design of math items, improve formative feedback systems, and enhance the reliability of automated assessment frameworks.




\section{Limitations}

As it is based on only three item models, the dataset is limited and needs to be scaled up to include both more item models and more instances of each model. We used a rather general rubric in this study; it is possible that a fine-grained rubric with more categories could uncover more insights about the nature of error patterns within solutions. In the interest of saving time, the LLM labeling and the human labeling were not independent; rather, the human verified the LLMs' labels for a portion of the data. Future work would examine agreement between LLM labeling and human labeling as independent processes. Nevertheless, LLM labeling with human verification reached 91.5\%  exact-match. Due to financial constraints, additional commercial models such as OpenAI o3 or more other LLM models like Anthropic Claude were not tested but could provide valuable insights and further evidence if included in future evaluations. 

\section{Conclusion}

This study systematically evaluated four LLMs, including two base models (OpenAI GPT-4o and DeepSeek-V3) and two advanced reasoning models (OpenAI o1 and DeepSeek-R1), across parallel arithmetic, algebraic, and number-theoretic item models in both single- and dual-agent paradigms. Models with stronger numerical competence, exemplified by o1, achieved step-level annotations that nearly doubled inter-rater agreement with human experts, underscoring their promise for scalable formative assessment. Dual-agent collaboration, mirroring the key benefits of human collaboration, further enhanced math problem-solving performance through cross-validation and emergent reasoning. In this study, the publicly released dataset, coding rubric, and benchmarking protocol equip researchers and practitioners with practical tools for pinpointing procedural versus conceptual breakdowns and for designing AI-enhanced teaching strategies. Future work will expand the problem bank, refine the error taxonomy, and integrate LLMs with external computational engines, bringing us closer to classroom-ready, pedagogically sound AI-based math instruction and assessment. 

\section*{Acknowledgments}
Thanks to Ikkyu Choi at ETS for generating the item model instances used in this research. Portions of the manuscript were refined with the assistance of AI tools solely for grammar checking and sentence refinement, without involvement in other study tasks. 


\bibliography{custom}




\end{document}